\documentclass[12pt]{article}

\usepackage[T2A]{fontenc}
\usepackage[utf8]{inputenc}
 
\usepackage{hyphenat}
\usepackage{xcolor}
\usepackage{hyperref}
\definecolor{linkcolor}{HTML}{000000}
\definecolor{citecolor}{HTML}{0B6B1A}
\definecolor{urlcolor}{HTML}{0B2A5B}
\hypersetup{pdfstartview=FitH, citecolor=citecolor, linkcolor=linkcolor, urlcolor=urlcolor, colorlinks=true}

\usepackage[pdftex]{graphicx}
\graphicspath{{.\\}}
\DeclareGraphicsExtensions{.pdf,.png,.jpg}

\usepackage{authblk}

\usepackage{array}
\usepackage{appendix}
\usepackage{amssymb}
\usepackage{float}
 
\begin{document}

\title{Investigating on RLHF methodology}
\author[ ]{Alexey Kutalev$^1$ and Sergei Markoff$^1$}
\affil[1]{SaluteDevices, Moscow, Russia}
\affil[ ]{\it kutalev@gmail.com}

 
\date{\today}


\maketitle

\begin{abstract}
 
In this article, we investigate the alignment of Large Language Models according to human preferences. We discuss the features of training a Preference Model, which simulates human preferences, and the methods and details we found essential for achieving the best results. 

We also discuss using Reinforcement Learning to fine-tune Large Language Models and describe the challenges we faced and the ways to overcome them. Additionally, we present our experience with the Direct Preference Optimization method, which enables us to align a Large Language Model with human preferences without creating a separate Preference Model. 

As our contribution, we introduce the approach for collecting a preference dataset through perplexity filtering, which makes the process of creating such a dataset for a specific Language Model much easier and more cost-effective.

\vspace{10pt} \textit{\textbf{Keywords:}
 Large Language Models, Reinforcement Learning from Human Feedback, Perplexity}

\end{abstract}

\newpage

\section{Introduction}

The launch of ChatGPT by OpenAI in late 2022 caused an explosive increase in interest in Large Language Models (LLMs) around the world. Many companies and research teams began developing their versions of instructive LLMs and conducting research on training, customizing, and using them to solve a wide range of tasks. 

The approach described in the paper \cite{instr-gpt} is often used as a starting point for these works. This approach includes three stages: pre-training a Language Model on a vast text corpus, fine-tuning it on a much smaller instructional dataset, and aligning the model according to human preference.

In case the target Language Model has a large size (in modern conditions, these are models with a size of 7 billion parameters or greater), the first stage of training requires huge expenditures of computing resources and, therefore, is available only to a narrow circle of organizations. The rest have to use ready-made pre-trained LLMs as a baseline. A neural network trained on this stage is able to generate a continuation for a given text.

Next, the model is trained further on an instructional dataset, and during this training the model learns to continue a given text by following the instructions embedded in it. The instructional dataset itself is a set of instructions and the results of their execution in natural language. This stage requires significantly less computing costs than the previous one and, therefore, is available to a wide range of enthusiasts. However, the quality of the resulting model strongly depends on the quality of the instructional dataset used for its training, all other things being equal. We will refer to the Language Model obtained from this stage as the Supervised Fine-Tuned model -- SFT.

The third stage is fine-tuning the model to align it with human preferences, that is, training the model to generate responses that are more suitable to the human criteria. This process usually requires creating an auxiliary Language Model that simulates human preferences, called a Preference Model. To create a Preferences Model, a preference dataset is needed, which is a set of examples where each example includes a user request and a pair of responses to this request. In these pairs, one of the responses is marked as more appropriate according to the human criteria, and another is marked as less appropriate. Most often, helpfulness and harmlessness are used as criteria for human-preferred responses. The markup of this dataset is usually done by human assessors \cite{ant}, but there are also examples of markup using other LLMs \cite{nectar}. It is worth noting an important fact separately: in all major papers on Reinforcement Learning from Human Feedback (e.g., \cite{instr-gpt}, \cite{ant}, \cite{llama2}), it is assumed that the preference dataset only includes responses generated by the model that is supposed to be aligned on the third stage. When the Preference model is created, the further training of the SFT model is performed using Reinforcement Learning (RL) with reinforcement from the Preference Model. Researchers mostly use the Proximal Policy Optimization (PPO, \cite{ppo}) method for this purpose.

There are alternative approaches to implement the third stage, such as Direct Preference Optimization (DPO, \cite{dpo}) or Odds Ratio Preference Optimization (ORPO, \cite{orpo}), where the target LLM is trained directly on the preference dataset, bypassing creating a separate Preference Model. 

This article focus on our experience in implementing the third stage.


\section{Modeling of human preferences}
\label{sec:modeling_hum_pref}


\subsection{Training the Preference Model}

Most papers on RLHF (\cite{fthp}, \cite{sfhf}, \cite{instr-gpt}, \cite{hf-rlhf}, etc.) use the following scheme of Preference model: in the basic LLM the final layer, called LM-head, is replaced by one-dimensional linear transformation. When we feed Preference Model with tokenized text of the request and response, this linear transformation translates the output hidden state for each token to a scalar value. The last one of these scalar values represents the preference score of the given response for the request. Thus, the Preference Model must work in such a way that for each request the preference score of more preferred response must be greater than the score for less preferred response. To indicate the end of response for the Preference Model a special token is usually added to the response text. This token is named "reward token" as the preference score is used in RL as reward value. 

The Preference Model faces a challenging task: it must differentiate between the responses generated by the LLM, determining which is better and which is worse by human criteria. Therefore, the Preference Model size should not be smaller but rather larger than the size of the LLM that generates the evaluated responses. In this scenario, the Preference Model acts as a discriminator, similar to how the discriminator in GAN networks operates. From the experience of training GANs (for example, \cite{gan} and \cite{gan1}), we know that a discriminator should be larger/smarter than the corresponding generator.

As mentioned above, when training a Preference Model, a dataset of requests and corresponding pairs of responses is used. Each pair has a markup indicating which response is better and which one is worse according to human preferences. The loss function used for this training typically has the following form:

\begin{equation}
\label{equ1}
loss(r_\theta) = - \mathbb{E}_{(x_i, y^b_i, y^w_i) \sim D} \log \sigma \left( r_\theta(x_i, y^b_i) - r_\theta(x_i, y^w_i) \right),
\end{equation}
where $x_i$ is $i$-th request, $y^b_i$ and $y^w_i$ are the better and worse responses for the $i$-th request in the preference dataset $D$, $r_\theta(x, y)$ is the scalar output (score) of Preference Model on request $x$ with response $y$.

\subsection{Important Features of the Preference Model}

It is necessary to highlight some crucial features of the Preference Model created according to the above scheme, which will be essential for further discussion.

Firstly, we have to note that the response scores given by the Preference Model for different requests may not be in a specific general range and may not be comparable. For example, for one user's request and the corresponding response generated by the Language Model, the Preference Model may assign a score of 1. For another user's request with a different response, the PM assigns a score of 100. However, this does not necessarily mean that the response to the second request is superior to the response to the first request. Instead, the response to the first request may be very good and has a score of 1, while the response to the second request may be mediocre and has a score of 100. It may happen due to the Preference Model training method, which only teaches PM to differentiate (rank) generated responses for a specific request rather than assign scores consistent across all requests and fitting into a specific rating range.

Secondly, the scale of the difference in the score values does not have to be the same for different requests. Consider that for a pair of generated responses to one request we have a difference in PM scores of 1000. And for a pair of responses to another request we have a difference in PM scores of 1. This situation does not mean that the difference between a pair of responses to the first request is more significant according to human preferences than the difference in a pair of responses to the second request.

Given the above, it seems unreasonable to us to use the difference in the scores from the Preference Model for a better response and a worse response to a given request as the "strength of differences" in the DPO training of the Language Model used by some authors, such as in \cite{ya-rlhf}.

\subsection{The scope of the Preference Model}

In the major papers on RLHF (\cite{instr-gpt}, \cite{ant}, \cite{llama2}), the authors state that the Preference Model (PM) performs well (i.e., it distinguishes well between better and worse responses) only within the range of generations of the basic Language Model, or more precisely, within the range of responses that were in the dataset for training Preference Model. And the ability of the PM to distinguish between better and worse responses according to human criteria degrades greatly when the distribution of responses generated by the LLM shifts. Such a shift can occur, for example, as a result of fine-tuning the LLM using PPO or DPO.

To overcome this problem, the preference dataset in \cite{instr-gpt}, \cite{ant}, and \cite{llama2} is constantly updated, and the Preference Model is trained further using examples with responses generated by the LLM, which was improved by the next PPO or DPO iteration.

In cases where the Preference Model is trained on the datasets with the responses that were not generated by the target LLM (for example, publicly available preference datasets such as Antropic Helpful, Antropic Harmless, Nectar, etc.), the range of practical applications of this Preference Model is obviously very different from the range of training examples in the preference dataset. Accordingly, we cannot expect this Preference Model to perform well on responses generated by target LLM.

For example, in the paper \cite{long-way-to-go}, the authors trained a Preference Model using public preference datasets and then used this PM to fine-tune their target LLM with PPO. As a result, they convincingly showed that the Preference Model did not learn any useful signal other than scoring the response length, excepting some special cases where they fine-tuned target LLM for code generation.

Our experience also confirms this. We initially attempted to train a Preference Model on a dataset compiled from requests and corresponding responses from various LLMs such as chatGPT 3.5, GPT-4, Llama and  Mistral of various versions, etc. We then evaluated the quality of the resulting Preference Model by counting the coincidence of the markup made by human assessors with the markup of the Preference Model on a dataset with pairs of responses generated by our target LLM. In this case, we obtained the accuracy at the level of 0.56, which is very close to random guessing.

\subsection{Comparison of Large Language Models}

In order to compare different LLMs, we conduct side-by-side comparisons based on the types of tasks given to the LLM. This approach allows us not only to obtain an average comparison, but also to understand on which types of tasks the LLM was improved or, conversely, worsened.

\begin{table}[h]
\centering
\begin{tabular}{ m{9em} m{22em} }
\textbf{Task type} & \textbf{Quality criteria for LLM responses}  \\
\multicolumn{2}{c}{ ~ } \\  \hline
Brainstorm & Helpfulness, Completeness, Truthfulness, Originality  \\ \hline
Chat & Empathy, Interestingness, Stylishness \\ \hline
Closed QA & Following the format for the response from the request, Truthfulness, Completeness, Argumentation \\ \hline
Generation & Interestingness, Stylishness \\ \hline
Open QA & Truthfulness, Completeness, Clarity, Conciseness  \\ \hline
Summarization & Completeness, Clarity, Conciseness, Stylishness \\ \hline
Rewrite & Completeness, Clarity, Conciseness, Stylishnes \\ \hline
Translation & Completeness, Stylishnes \\ \hline
Classification & Following the format for the response from the request, Completeness, Accuracy  \\ \hline
Extract & Following the format for the response from the request, Completeness, Conciseness \\ \hline

\end{tabular}
\caption{Quality criteria depending on the task type. We listed the criteria in descending order of importance. For all task types we also used the criteria for accurately following the user's instructions from the request and harmlessness while comparing the responses.}
\label{table:1}
\end{table}

At the beginning of our experiments with RLHF, we noticed that when using a general preference dataset, the side-by-side evaluation between the basic LLM (SFT) and the model trained by RLHF methods is very uneven by task type. Therefore, we divided the preference dataset by task type and marked it up using separate response quality criteria for each type. For example, for the "Chat" type of task, a better response is more interesting, empathetic, and engaging in conversation, while it does not have to be truthful, complete, or concise. For the "Open QA" type of task, а better response must be more truthful and complete than a worse one. The full set of criteria for each task can be found in Table \ref{table:1}.

\begin{table}[h]
\centering
\begin{tabular}{ m{12.5em} m{12.5em} }
\textbf{Task type} & \textbf{\% examples where a better response is longer}  \\
\multicolumn{2}{c}{ ~ } \\ \hline
Brainstorm &  74.13\%  \\ \hline
Cat &  76.57\% \\ \hline
Closed QA &  67.87\%  \\ \hline
Generation &  65.26\%  \\ \hline
Open QA &  73.23\%  \\ \hline
Summarization &  77.95\% \\ \hline
Rewrite &  71.54\% \\ \hline
Translation &  69.87\% \\ \hline
Classification &  65.69\%  \\ \hline
Extract &  55.58\% \\ \hline

\end{tabular}
\caption{Proportions in our preference dataset between the number of examples where a better response is longer and the number of examples where a better response is shorter. The proportions are shown for all task types.}
\label{table:2}
\end{table}

In addition, we found the strong correlation between the quality of the response and its length. That is, longer responses turned out to be better in more cases than shorter responses for most types of tasks. The proportion in our preference dataset of the better-response-longer and better-response-shorter examples is shown in Table \ref{table:2} by task type.

Taking into account the results of \cite{long-way-to-go}, we tried to understand whether our Preference Model was able to learn how to correctly rank responses to a request based on criteria other than length. To do this, we trained a Preference Model using a dataset that was balanced by the number of examples where a better response is longer and a better response is shorter, and we also trained a Preference Model without such balancing. In this case, the measure of success is PM ranking accuracy growth on the test subset of the preference dataset, divided into two parts according to the criterion "a better response is longer" and "a better response is shorter". The accuracy measurement results are presented in the Table \ref{table:3}.

\begin{table}[h]
\centering
\begin{tabular}{ m{9em} m{10em} m{10em} }
\textbf{Task type} & \textbf{Accuracy of the PM ranking on the test subset where a better response is longer} & \textbf{Accuracy of the PM ranking on the test subset where a better response is shorter} \\
\multicolumn{3}{c}{ ~ } \\ \hline
Brainstorm & 0.782 & 0.577 \\ \hline
Chat & 0.779 & 0.691 \\ \hline
Closed QA & 0.864 & 0.824 \\ \hline
Generation & 0.765 & 0.780 \\ \hline
Open QA & 0.785 & 0.688 \\ \hline
Summarization & 0.781 & 0.594 \\ \hline
Rewrite & 0.741 & 0.484 \\ \hline
Translation & 0.900 & 0.781 \\ \hline
Classification &  0.797 & 0.563 \\ \hline
Extract & 0.768 & 0.744 \\ \hline

\end{tabular}
\caption{Evaluation results of the Preference Model, which was trained on our preference dataset that we balanced by the number of better-response-longer and better-response-shorter examples. The evaluation was done on test subsets for all task types.}
\label{table:3}
\end{table}

From the results in Table \ref{table:3}, we can see that the accuracy for almost all types of tasks is significantly different from 0.5, i.e. from random guessing. This result indicates that during training our Preference Model it was possible to extract from the preference dataset some criteria for distinguishing between better and worse responses other than just their length.

To understand how well our Preference Model performs in the target area, we measured the agreement between the markup produced by the Preference Model and the markup produced by human assessors using the examples with the responses generated by our basic LLM (SFT). The results of this analysis can be found in Table \ref{table:4}. We can see that, for most types of tasks, the Preference Model accuracy on an evaluative subset with "better-response-shorter" examples is comparable to or less than 0.5, which is equivalent to or worse than random guessing. This proves that our PM training is unsatisfactory at extracting criteria for ranking responses (other than response length) from the preference dataset, confirming the conclusions made in \cite{long-way-to-go}.

\begin{table}[h]
\centering
\begin{tabular}{ m{9em} m{10em} m{10em} }
\textbf{Task type} & \textbf{Accuracy of the PM ranking on the test subset where a better response is longer} & \textbf{Accuracy of the PM ranking on the test subset where a better response is shorter} \\
\multicolumn{3}{c}{ ~ } \\ \hline
Brainstorm & 0.588 & 0.452 \\ \hline
Chat & 0.649 & 0.444 \\ \hline
Closed QA & 0.613 & 0.567 \\ \hline
Generation & 0.653 & 0.618 \\ \hline
Open QA & 0.538 & 0.500 \\ \hline
Summarization & 0.717 & 0.337 \\ \hline
Rewrite & 0.714 & 0.455 \\ \hline
Translation & 0.722 & 0.365 \\ \hline
Classification &  0.678 & 0.456 \\ \hline
Extract & 0.746 & 0.364 \\ \hline

\end{tabular}
\caption{Evaluation of the Preference Model quality based on a special test dataset. The examples in this dataset include pairs of responses generated by the basic LLM (SFT) and marked up by our human assessors.}
\label{table:4}
\end{table}

Note once again that all the above evaluations are made for a Preference Model that was trained on a dataset consisting of responses that were not generated by the LLM, which was used as the baseline for this Preference Model.

\subsection{Features of the implementation of our Preference Model}

In our experiments, we followed the above architecture and the training method of the Preference Model. We used our LLM (SFT) without the LM-head as a baseline and added a linear layer to it. This layer converts the output hidden state of each token into a score value. Thus, the PM score of a response to a request is taken from the last token position of the tokenized request-response text.

We tried to train the Preference Model in two different ways. The first way involved training only the LoRA adapters to dense matrices of the basic model and the output linear layer. In this approach, we set the rank of the LoRA adapters to 128, and this significantly reduced GPU memory consumption but still allowed for sufficient changes to the model. The second way involved training the entire model. In this case, we first trained the output linear layer while keeping the remaining parameters frozen. We performed this preliminary training because our experiments showed that training the entire model at once with an added untrained output layer worsens the model at the initial stage of training.

Our experiments did not show any significant difference in the quality of the resultant Preference Model when training entire model and training only LoRA-adapters and output linear layer. Therefore, we decided to use the LoRA-adapters option as it is more memory-efficient.

\section{Reinforcement Learning using PPO}

\subsection{Implementation difficulties}

In order to implement reinforcement learning for our basic LLM (SFT) in correspondence with the scheme described in \cite{instr-gpt}, \cite{ant} and \cite{hf-rlhf}, several difficulties must be overcome.

First, during the operation of the RLHF-PPO algorithm, 4 models must be available simultaneously: the target LLM, the basic LLM (SFT), the Сritic model (predicting a cumulative future reward for each response token), and the Preference Model. The target LLM and the Critic model must be trainable, the target model must also generate responses at each PPO iteration. The basic LLM (SFT) and the Preference Model must work in the inference mode as they only perform forward passes and are not trained.

In order to fit all 4 models into the memory of our NVIDIA A100 GPU cluster, we applied a scheme where only the basic LLM (SFT) is loaded into the GPU memory, and the other three models are represented as sets of LoRA adapters and output linear layers (heads) to the basic model. We used the DeepSpeed framework for training with ZeRO2 mode for 7b models and ZeRO3 for 29b models.

Second, the authors of \cite{dpo} and \cite{rlhf-sec-1} report that they faced the instability of the PPO algorithm during the learning process when applying the classical PPO approach from \cite{ppo} to train LLMs. This instability is due to significant fluctuations in the average value and range of the scores (rewards) which Preference Model provides. 

In order to stabilize the PPO algorithm, the authors of \cite{rlhf-sec-1} proposed to normalize the scores from a Preference Model using moving average and moving standard deviation. However, in our experiments, this scheme did not produce the desired results, that is, the algorithm often diverged suddenly.

We obtained much better results by normalizing the advantage values in each batch of training examples, similar to the implementation in the TRL and TRLX packages. Due to the structure of the loss function in PPO, this normalization leads to the fact that the norm of the loss function gradient remains approximately the same throughout the training period. As a result, it is possible to select the training parameters for a specific LLM so that the learning process remains stable for a long time.

However, such a scheme has a drawback: when the growth of the scores given by the Preference Model for generated responses begins to slow down or stop completely, the norm of the loss function gradient does not decrease but remains approximately constant, while the gradient itself becomes chaotic. That leads to the gradual deterioration of the trained LLM at a later stage of PPO training. In this case, it is necessary to use heuristics to stop training early or decrease the learning rate when a significant increase in the average PM score of responses has already been obtained, but the integrity of the trained LLM is still preserved.

\subsection{Results}

We performed training of our basic LLM (SFT) using the PPO algorithm on the dataset of requests balanced by the number of requests for each task type.

\begin{figure}[h]
\centering
\includegraphics[width=1\textwidth]{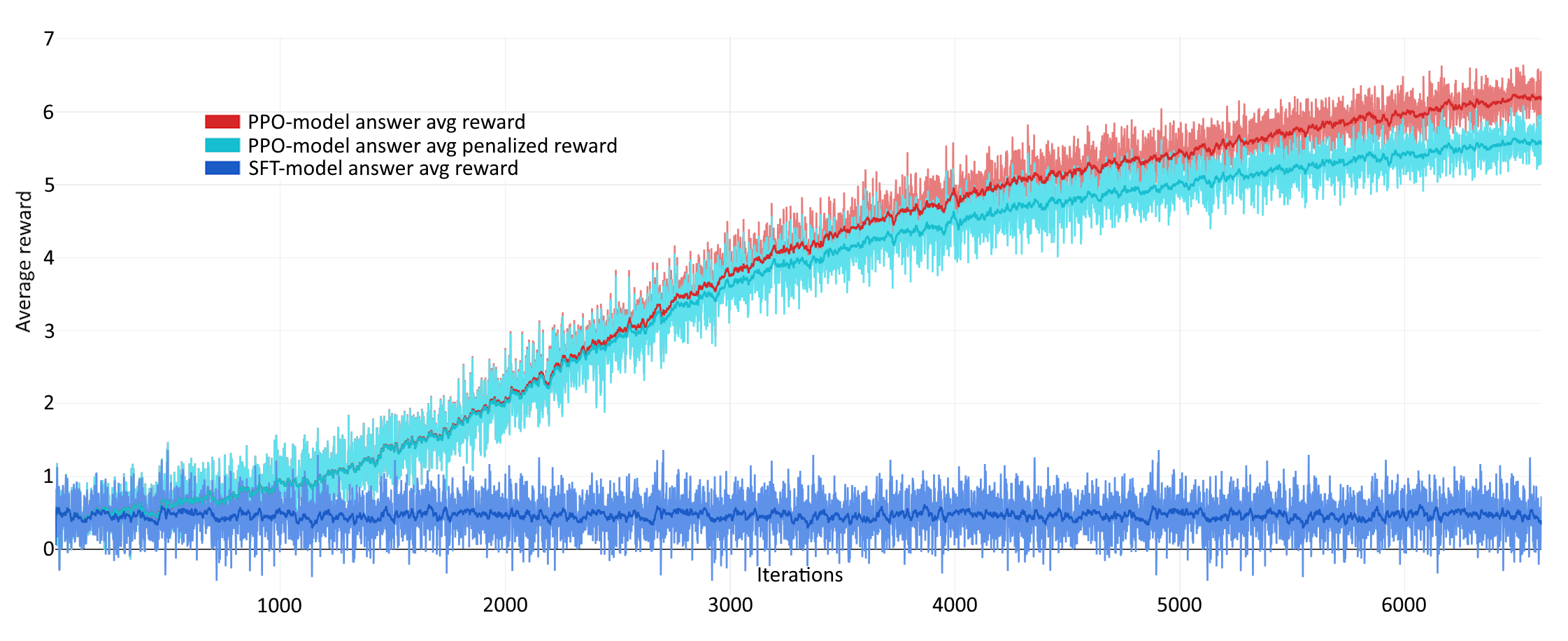}
\caption{Average PM score for the responses generated by the trained LLM during the PPO training with both the plain average score and the average score minus the KLD-penalty for deviation from the basic LLM distribution. Additionally, the graph shows the average score for the responses from basic LLM for the same requests.}
\label{figure:1}
\end{figure}

In our experiments, we achieved a steady, consistent growth in the average score of responses generated by trained LLM. As a result of  PPO training, we obtained the model that generates the responses with PM scores significantly higher than PM scores for the responses generated by the basic LLM (SFT) for the same requests. The graph in Figure \ref{figure:1} shows the growth dynamics in the average PM score of responses generated by the trained model in the process of PPO.

\begin{figure}[H]
\centering
\includegraphics[width=1\textwidth]{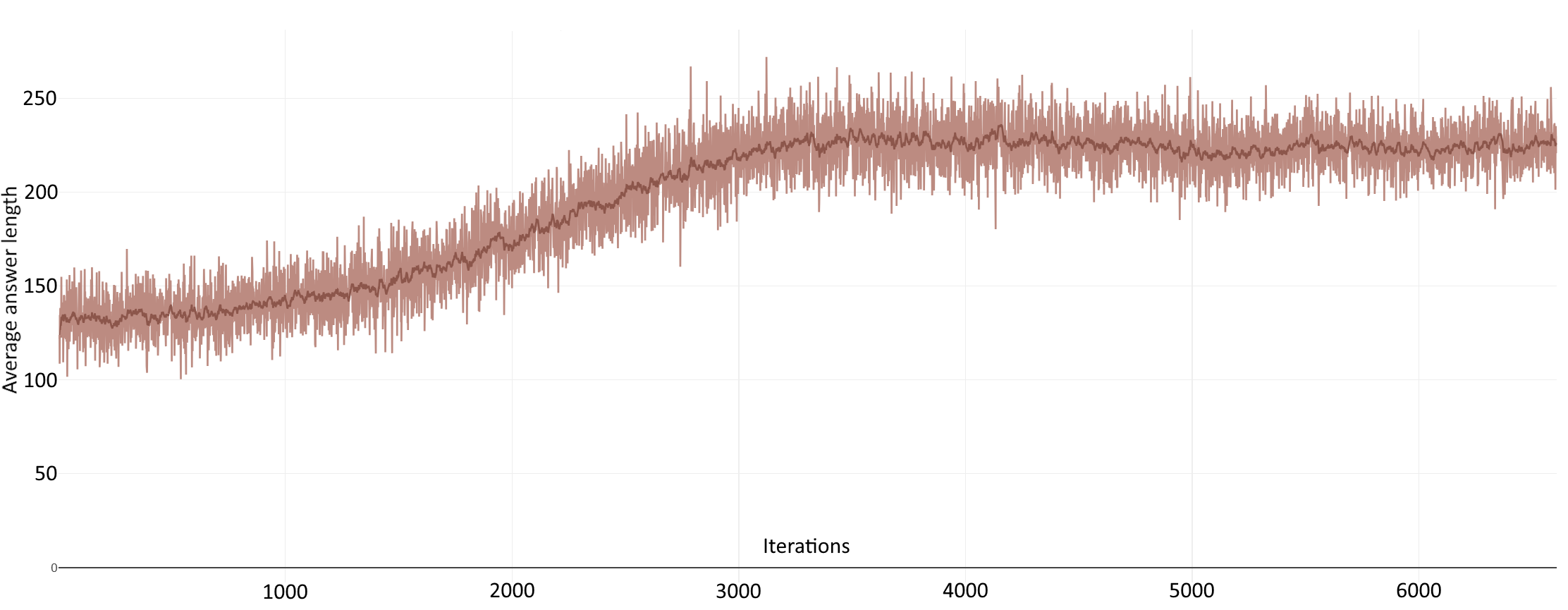}
\caption{Average length of response generated by the trained LLM during the PPO process.}
\label{figure:2}
\end{figure}

\begin{figure}[H]
\centering
\includegraphics[width=1\textwidth]{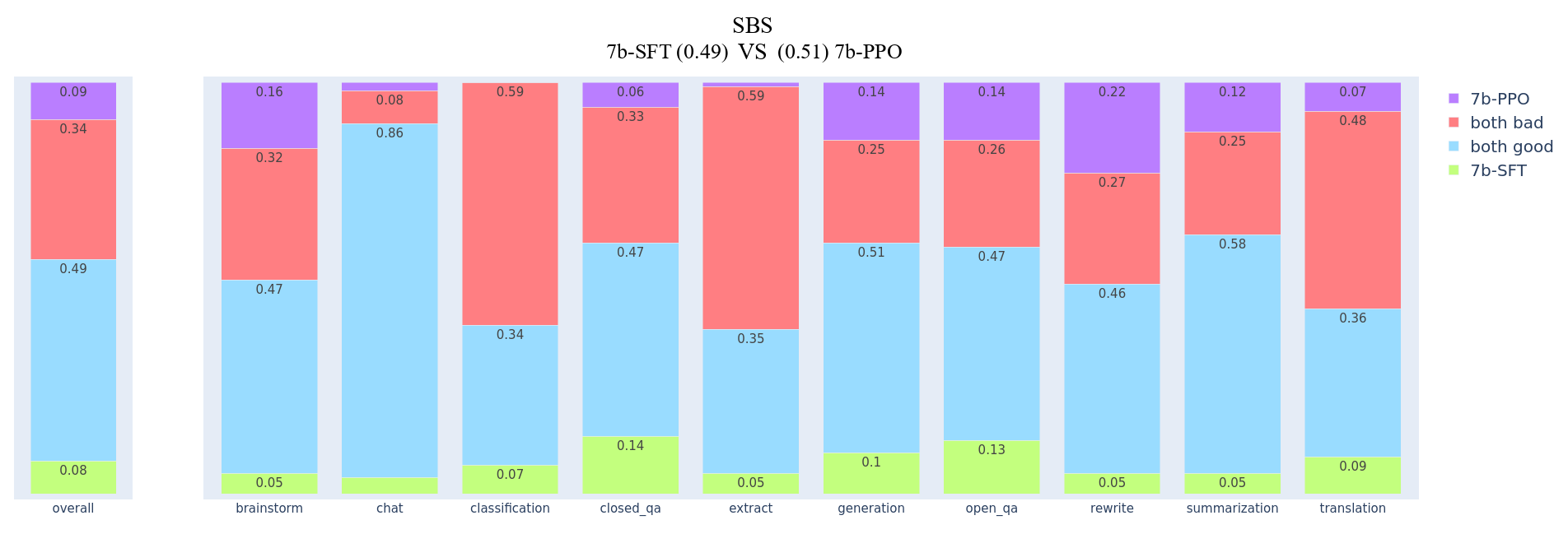}
\caption{Side-by-side evaluation of the basic LLM (7b-SFT) versus the LLM trained with PPO (7b-PPO).}
\label{figure:3}
\end{figure}

At the same time, during the PPO training process, we evaluated the trained LLM using several well-known LLM evaluation metrics, such as GSM8k, HUMAN-EVAL, MMLU, RUBQ, WINOGRANDE, etc. If any of the metrics began to deteriorate, we used this fact as a criterion for an early stop of learning. As a result, these metrics did not deteriorate or deteriorated insignificantly for the final trained LLM.

We also noticed that the average response length was increasing consistently together with the average PM score growth at the initial stage of PPO training. However, later, the growth in the average PM score started to occur due to other changes in the LLM, as seen in Figure \ref{figure:2}.

As shown in Figure \ref{figure:3}, the side-by-side comparison of the final trained LLM with the basic LLM (SFT) indicates that the model's performance was improved for some task types but worsened for others. This could be because, within the range of the basic model generation, the scoring of responses by the Preference Model is less effective for responses to requests from these worsened task types, as described in the section about the Preference Model above.

\section{Training LLM using the DPO algorithm}

\subsection{Brief description of the DPO}

In the paper \cite{dpo}, the authors proposed the method of direct training of the basic LLM (SFT) using a preference dataset, bypassing the creation of a separate Preference Model.

Its essence is to train a LLM using a special kind of loss function (see formula \ref{equ2}), which leads to an increase in the probability of generating more preferred responses and a decrease in the probability of generating less preferred responses. At the same time, the loss function includes a penalty for significant deviation from the basic LLM. This penalty is based on the difference in distributions generated by the trained LLM from the distributions of the basic LLM, and it protects the trained LLM from destructive changes.
$$
loss(\pi_{DPO}, \pi_{SFT}) =
$$
\begin{equation}
\label{equ2}
 - \mathbb{E}_{(x_i, y^b_i, y^w_i) \sim D} \left[ \log \sigma \left( \beta \log \frac{\pi_{DPO}(y^b_i | x_i)}{\pi_{SFT}(y^b_i | x_i)}   -  \beta \log \frac{\pi_{DPO}(y^w_i | x_i)}{\pi_{SFT}(y^w_i | x_i)} \right) \right],
\end{equation}
where $x_i$ is $i$-th request, $y^b_i$ is better response, $y^w_i$ is worse response for the $i$-th request in the preference dataset $D$, $\pi_{DPO}(y|x)$ is likelihood of response $y$ when requesting $x$ on the LLM trained with DPO, $\pi_{SFT}(y|x)$ is likelihood of response $y$ when requesting $x$ on the basic LLM (SFT), $\beta$ is the penalty coefficient to prevent the distribution from deviating to far from the basic model.

It is worth noting that the authors of the paper \cite{dpo} explicitly indicate that for successful application of DPO, the preference dataset must consist of responses generated by the basic LLM, that is, it must be on-distribution for this LLM.

\subsection{The ability to use the out-of-distribution dataset for DPO training}

The preference dataset we collected (approximately 200,000 examples) is not on-distribution for our basic models 7b-SFT and 29b-SFT. Nevertheless, we tried to train our basic 7b LLM on this dataset using DPO. 

Similar to the PPO case, to select the parameters of DPO training and control the results, we were measuring a set of well-known metrics, such as GSM8k, HUMAN-EVAL, MMLU, RUBQ, and WINOGRANDE. Next, we were selecting only those DPO-trained models as candidates for side-by-side evaluation whose metrics did not deteriorate.

As a result of the above, we conducted DPO training during 1 epoch with a learning rate of $1 \cdot 10^{-7}$ and a batch size of 256.

\begin{figure}[h]
\centering
\includegraphics[width=1\textwidth]{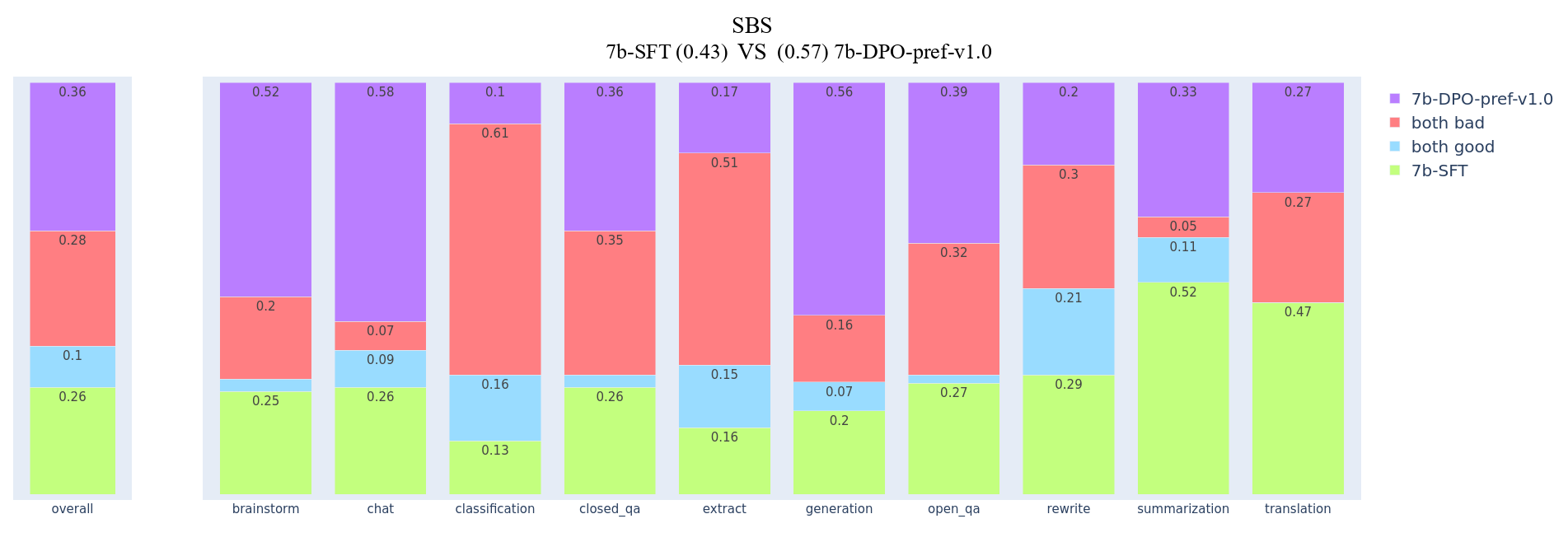}
\caption{Side-by-side evaluation of the basic LLM (7b-SFT) versus the LLM trained with DPO on our preference dataset (7b-DPO-pref-v1.0).}
\label{figure:4}
\end{figure}

In Figure \ref{figure:4}, we present the results of a side-by-side comparison of the final 7b-DPO model with the basic 7b-SFT model. Despite using an out-of-distribution dataset for training, the DPO model has improved according to the average estimation, with a DPO-win of 57 compared to SFT-win of 43. However, we can also see that the model's performance on certain tasks has deteriorated, which we believe is an unacceptable price for the overall improvement.

\subsection{Creating an on-distribution preference dataset for DPO using filtering by perplexity}

The development of the language corpus used for pre-training our LLM and the improvement of the instructional dataset used for training the LLM to follow instructions have never stopped. Therefore, we continuously obtain new versions of our basic SFT models. To have an on-distribution preference dataset for RLHF purposes, we need to compose this dataset entirely from responses generated by our current SFT model and then mark it up by human assessors. In these conditions, maintaining an up-to-date, large-scale preference dataset that is on-distribution for each new version of our SFT model becomes a very expensive and difficult task.

To address this challenge, we developed the method of creating an on-distribution preference dataset by filtering examples from a union of available preference datasets, including publicly available ones, based on perplexity. The method's main idea is to select examples from a union of available preference datasets, where both responses have low perplexity on the current basic LLM. Suppose both responses in the preference example have perplexity that is in the perplexity range of the responses generated by the current basic LLM. So, we can consider such an example as on-distribution for this LLM. Further, we can use the subset of examples selected in this way to train the current basic LLM using DPO.

Accordingly, to create such a preference dataset, we followed these steps:

\begin{enumerate} 
\item We generated responses to a certain set of requests for each type of task using our basic 7b-SFT LLM and calculated the perplexity for each response. Then, for each type of task, we calculated the 95-percentile of all perplexities of generated responses. We did not take a simple maximum to eliminate outliers, that is, random unlikely generations. The obtained 95-percentile for each task type will set further the boundary for filtering.

\item We merged our preference dataset and a very large publicly available dataset Nectar from \cite{nectar}. For each example in the combined preference dataset, we calculated the perplexity of each response on our basic 7b-SFT LLM.

\item From all examples of the combined dataset we selected only those examples where the perplexity of both responses was less than the calculated border for the task type of this example.

\item We also balanced the resulting dataset based on task types so that the sizes of the subsets for different task types differed by no more than 2 times.  

\end{enumerate}

Finally, for our 7b-SFT model, we have a preference dataset with a size of about 380,000 examples. 

Similarly to the previous section, we conducted the selection of hyperparameters for DPO training and the selection of candidates for side-by-side evaluation based on the non-deterioration of a set of metrics for LLMs. 

As a result, we conducted DPO training during 2 epochs with a learning rate of $1 \cdot 10^{-8}$ and a batch size of 256.

\begin{figure}[h]
\centering
\includegraphics[width=1\textwidth]{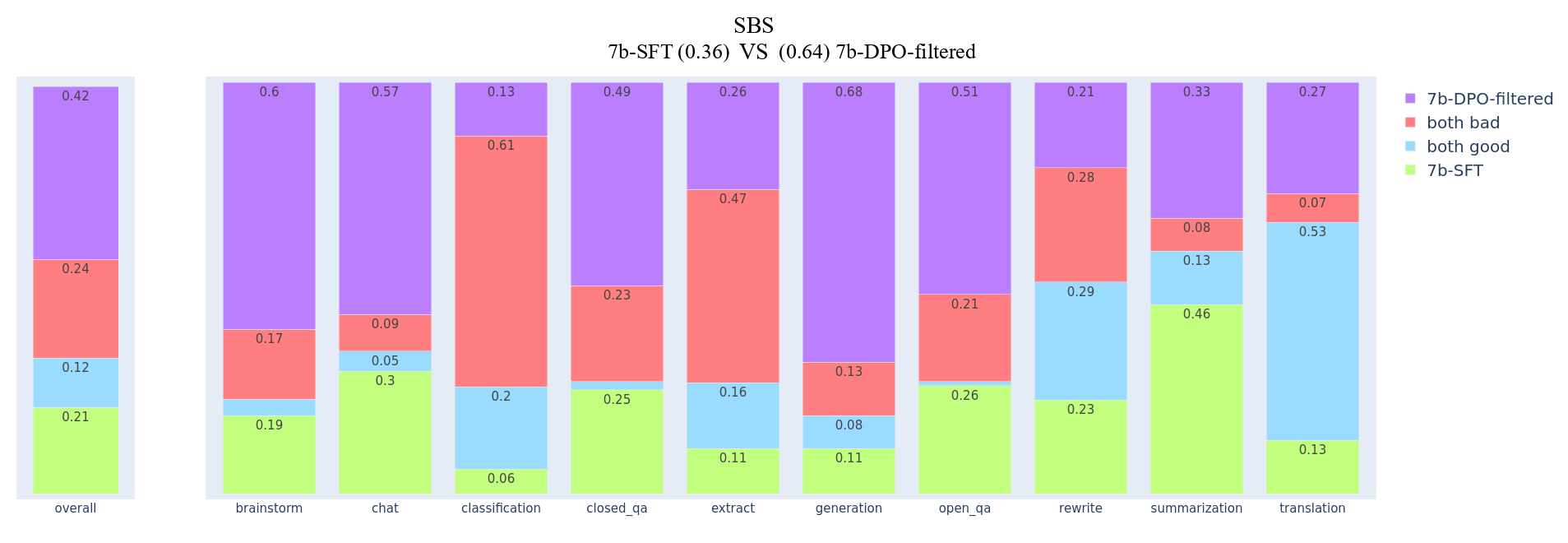}
\caption{Side-by-side comparison between the basic LLM (7b-SFT) and the LLM trained using DPO on the preference dataset that was created by filtering based on perplexity (7b-DPO-filtered).}
\label{figure:5}
\end{figure}

Figure \ref{figure:5} shows the results of side-by-side evaluation of the model trained by DPO against the basic SFT model. In this case, the situation is different for the better. The DPO model wins with a significantly higher total score - DPO-win 64 versus SFT-win 36. There are small deteriorations only on 2 task types out of 10. This slight deterioration in the model's abilities on 2 task types seems to be an acceptable price for a large average improvement and improvements on the remaining 8 types.

\subsection{Performing DPO training using the preference dataset, which includes only responses generated by the model}

As a proof of concept of creating a preference dataset from the papers \cite{instr-gpt} and \cite{ant}, we created a small preference dataset where the examples include only responses generated by the basic LLM, that is, a model that will then have to be trained using DPO on this dataset.

To do this, we collected a set of requests for each of the 10 types of tasks and generated several responses to each request using our basic 7b-SFT model. Then, for each request, we selected 2 most different responses and created an example with these request and responses. The resulting set of examples (request with two responses) were given to human assessors for markup to select the better and worse response in the example.

The markup was performed using an overlap of 3, and after that, we selected only those examples that received unambiguous and consistent markup from all markup assessors. As a result, we obtained a strictly on-distribution preference dataset of about 11,000 examples.

We selected hyperparameters in the same way as in previous cases, and then we conducted DPO training for our 7b-SFT model with the above small dataset. After that we selected the optimal candidate for side-by-side evaluation. The training was performed over 6 epochs with a learning rate of $1 \cdot 10^{-7}$ and a batch size of 256.

\begin{figure}[h]
\centering
\includegraphics[width=1\textwidth]{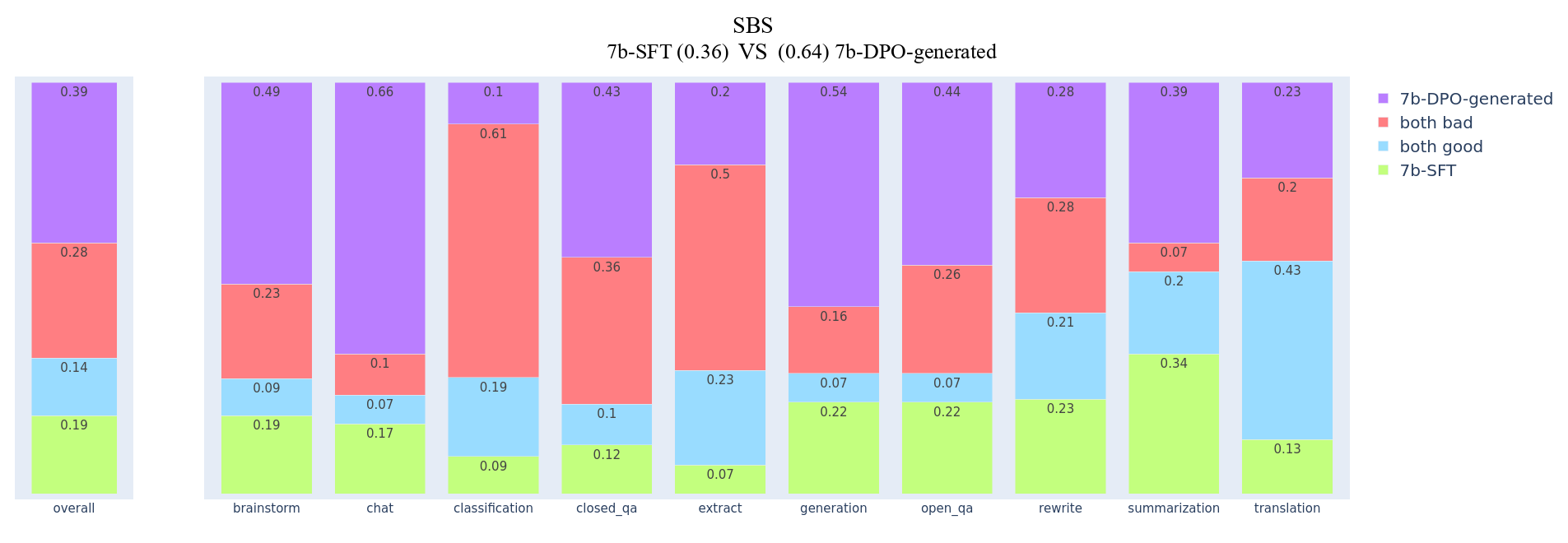}
\caption{Side-by-side comparison of the basic LLM (7b-SFT) with the LLM trained using DPO (7b-DPO-generated) on a preference dataset with responses generated by the basic model.}
\label{figure:6}
\end{figure}

Figure \ref{figure:6} shows the results of a side-by-side evaluation of the  model trained by DPO against the basic SFT model. The DPO model wins with a large total score --- DPO-win 64 against SFT-win 36, as well as in the case of using perplexity filtering. But in this case there is no deterioration in any of the 10 types of tasks.

\subsection{Using DPO training to overcome the problem of repeats when LLM generates the response}

Following the method described in \cite{deepseek}, we applied DPO training to overcome multiple repeats of text piece in the responses of our 7b LLM on some requests.

To do this, from our archive of requests and their responses generated by our LLMs, we selected requests whose responses have repetitions (multiple repetitions --- a substring with a length of more than 20 characters is repeated 7 or more times in the response, tandem repetitions --- a substring with a length of more than 100 characters is repeated 2 or more times together) or cycles (when the model generates repeating text until the length of the LLM context).

For a set of these requests, we also generated responses from our LLMs without repetitions and cycles. Then, we created the preference dataset where the responses without repetitions and cycles were marked as more preferred (better), while responses with repetitions or cycles were marked as less preferred (worse) for corresponding requests. Thus, we obtained a preference dataset for overcoming repeats with about 40,000 examples.

We selected only on-distribution examples from this dataset for our basic 7b-SFT model using the filtering by perplexity method described above. It turned out about 7,000 examples.

Having selected the training parameters as before, we conducted DPO training of our basic 7b-SFT LLM on this dataset with 7,000 examples. To evaluate the results, we used a special set of requests in the responses to which our models are prone to produce repetitions and cycles. As a result, the number of responses with repetitions or cycles decreased by 3 or more times on this set of requests --- see table \ref{table:5}.

\begin{table}[h]
\centering
\begin{tabular}{ m{8em} m{9em} m{9em} }
\textbf{Model} & \textbf{\% cycles} & \textbf{\% repetitions} \\
\multicolumn{3}{c}{ ~ } \\ \hline
7b-SFT & 2.19\% & 11.7\% \\ \hline
7b-DPO & 0.42\% & 3.55\% \\ \hline
\end{tabular}
\caption{Percentage of repetitions and cycles among the responses generated by our basic 7b-SFT model and trained 7b-DPO model on a special test set of requests, where our models tend to generate responses with repetitions or cycles.}
\label{table:5}
\end{table}

Similarly, using filtering by perplexity, we selected on-distribution examples for our larger model 29b-SFT --- about 6,000 examples. After conducting DPO training of the 29b-SFT model on this dataset, we also got a significant reduction in repetitions and cycles, as shown in Table \ref{table:6}.

\begin{table}[h]
\centering
\begin{tabular}{ m{8em} m{9em} m{9em} } 
\textbf{Model} & \textbf{\% cycles} & \textbf{\% repetitions} \\
\multicolumn{3}{c}{ ~ } \\ \hline
29b-SFT & 9.41\% & 20.43\% \\ \hline
29b-DPO & 2.42\% & 8.71\% \\ \hline
\end{tabular}
\caption{Percentage of repetitions and cycles among the responses generated by our basic 29b-SFT model and trained 29b-DPO model on a special test set of requests, where our models tend to generate responses with repetitions or cycles.}
\label{table:6}
\end{table}

\section{Conclusions, discussions, and hypotheses about the formation of a preference dataset}

As mentioned above in the section about the DPO method, formation and maintaining a relevant preference dataset of the required size composed entirely of responses generated by the current basic model and marked up by human assessors is a costly and difficult task. In addition, the markup of the preference dataset, carried out even by professional assessors, often contains ambiguous choices, and sometimes just errors.

As shown by the researchers from the paper \cite{rlhf-sec-2}, who performed the markup of the preference dataset using the Preference Model trained on it, publicly available preference datasets contain many examples where the markup directly contradicts to the markup performed by the Preference Model (i.e., the difference in scores of better and worse responses is significantly negative). Our experience of training the Preference Model, and then using it for checking the markup it our preference dataset, revealed a similar situation. Many examples where the PM score of the worse response is greater than the PM score of the better one, upon detailed examination, really turned out to be incorrectly marked by human assessors.

Considering all of the above, it would be good to reduce the cost of the markup task and make it more accurate and reliable. To address this issue, we have explored several different approaches:

\begin{itemize}

\item We attempted to automate the process of marking up LLM responses by transferring the selection of the preferred response to the LLM itself using a specific type of request. We tried to use our 29b-SFT model and GPT4 for this purpose. Unfortunately, this approach has not yet brought positive results. We performed two markups of the test preference dataset: one by LLM and one by human assessors. The matching between these two markups was about 50\%, which is equivalent to random guessing.

\item Following the method from \cite{self-refine}, we also tried to ask our basic LLM to generate a more preferable response to the request based on a given request-response pair. However, human markup on the obtained pairs of responses showed that only about 50\% of the model's responses were improved.

\item In order to create a preference dataset that would be on-distribution for our current basic LLM, we also tried to apply a perplexity filtering method, which we developed inspired by \cite{ppl}. After collecting a lot of responses generated by our current basic LLM, we determined the maximum threshold --- such a number that most (95\%) of the generated responses have a perplexity less than this threshold. Then, we merged several publicly available preference datasets with our proprietary preference dataset into one set of examples. After that, we selected from this merged dataset only those examples where both responses have a perplexity on our current basic LLM below the calculated threshold (that is, in theory, these responses could be generated by our current basic LLM). After that, we performed DPO training our current basic LLM on the subset of examples selected in this way. We have described the corresponding results above in the section about DPO.

\end{itemize}

Only the last one of these three approaches showed good results, that is, it allowed us to create such a preference dataset for our 7b-SFT model, which gave it a significant improvement within RLHF training.

We hope that the proposed approach of creating a preference dataset by filtering a large array of data by perplexity will help to popularize RLHF methods since this approach avoids the most expensive and difficult part of the process: marking up a preference dataset by human assessors.




\newpage

\begin{thebibliography}{99}

\bibitem{ppo} John Schulman et al (2017) Proximal policy optimization algorithms. \textit{arXiv}:1707.06347. DOI: 10.48550/arXiv.1707.06347.

\bibitem{fthp} Daniel M. Ziegler et al (2019) Fine-Tuning Language Models from Human Preferences. \textit{arXiv}:1909.08593. DOI: 10.48550/arXiv.1909.08593.

\bibitem{sfhf} Nisan Stiennon et al (2020) Learning to summarize from human feedback. \textit{arXiv}:2009.01325. DOI: 10.48550/arXiv.2009.01325.

\bibitem{instr-gpt} Long Ouyang et al (2022) Training language models to follow instructions with human feedback. \textit{arXiv}:2203.02155. DOI: 10.48550/arXiv.2203.02155.

\bibitem{ant} Yuntao Bai et al (2022) Training a Helpful and Harmless Assistant with Reinforcement Learning from Human Feedback. \textit{arXiv}:2204.05862. DOI: 10.48550/arXiv.2204.05862.

\bibitem{hf-rlhf} Nathan Lambert et al (2022) Illustrating Reinforcement Learning from Human Feedback (RLHF). \textit{Hugging Face Blog}.
\href{https://huggingface.co/blog/rlhf}{https://huggingface.co/blog/rlhf}.

\bibitem{llama2} Hugo Touvron et al (2023) Llama 2: Open Foundation and Fine-Tuned Chat Models. \textit{arXiv}:2307.09288. DOI: 10.48550/arXiv.2307.09288.

\bibitem{deepseek} Xiao Bi et al (2024) DeepSeek LLM: Scaling Open-Source Language Models with Longtermism. \textit{arXiv}:2401.02954. DOI: 10.48550/arXiv.2401.02954.

\bibitem{rlhf-sec-1} Rui Zheng et al (2023) Secrets of RLHF in Large Language Models Part I: PPO. \textit{arXiv}:2307.04964. DOI: 10.48550/arXiv.2307.04964.

\bibitem{rlhf-sec-2} Binghai Wang et al (2024) Secrets of RLHF in Large Language Models Part II: Reward Modeling. \textit{arXiv}:2401.06080. DOI: 10.48550/arXiv.2401.06080.

\bibitem{nectar} Banghua Zhu et al (2022) Nectar Dataset. \textit{Hugging Face Dataset Cards}.
\href{https://huggingface.co/datasets/berkeley-nest/Nectar}{https://huggingface.co/datasets/berkeley-nest/Nectar}.

\bibitem{gan} Ian J. Goodfellow et al (2014) Generative Adversarial Networks. \textit{arXiv}:1406.2661. DOI: 10.48550/arXiv.1406.2661.

\bibitem{gan1} Antonia Creswell et al (2017) Generative Adversarial Networks: An Overview. \textit{arXiv}:1710.07035. DOI: 10.48550/arXiv.1710.07035.

\bibitem{ya-rlhf} Zinov N. (2023) YandexGPT: alignment LLM in details. \textit{AI Jorney}. \href{https://vk.com/video-22522055_456243329}{Report}.


\bibitem{long-way-to-go} Prasann Singhal et al (2023) A Long Way to Go: Investigating Length Correlations in RLHF. \textit{arXiv}:2310.03716. DOI: 10.48550/arXiv.2310.03716.

\bibitem{dpo} Rafael Rafailov et al (2023) Direct Preference Optimization: Your Language Model is Secretly a Reward Model. \textit{arXiv}:2305.18290. DOI: 10.48550/arXiv.2305.18290.

\bibitem{orpo} Jiwoo Hong et al (2024) ORPO: Monolithic Preference Optimization without Reference Model. \textit{arXiv}:2403.07691. DOI: 10.48550/arXiv.2403.07691.

\bibitem{self-refine} Aman Madaan et al (2023) Self-Refine: Iterative Refinement with Self-Feedback. \textit{arXiv}:2303.17651. DOI: 10.48550/arXiv.2303.17651.

\bibitem{ppl} Zachary Ankner et al (2024) Perplexed by Perplexity: Perplexity-Based Data Pruning With Small Reference Models. \textit{arXiv}:2405.20541. DOI: 10.48550/arXiv.2405.20541.



\end{thebibliography}
\end{document}